\documentclass[conference]{IEEEtran}

\usepackage{amsmath,amssymb}
\usepackage{graphicx}
\usepackage{booktabs}
\usepackage{hyperref}
\usepackage{xcolor}
\usepackage{algorithm}
\usepackage{algorithmic}
\usepackage{enumitem}
\usepackage{subcaption}

\title{Learning to Trust the Crowd: A Multi-Model Consensus Reasoning Engine for Large Language Models}

\author{
\IEEEauthorblockN{Pranav Kallem}
\IEEEauthorblockA{Department of Computer Science\\
The University of Texas at Austin\\
Austin, TX, USA}
}

\begin{document}
\maketitle

\begin{abstract}
Large language models (LLMs) achieve strong average performance yet remain unreliable at the instance level, with frequent hallucinations, brittle failures, and poorly calibrated confidence. We study reliability through the lens of \emph{multi-model consensus}: given responses from several heterogeneous LLMs, can we learn which answer is most likely correct for a given query? We introduce a \emph{Multi-Model Consensus Reasoning Engine} that treats the set of LLM outputs as input to a supervised meta-learner. The system maps natural language responses into structured features using semantic embeddings, pairwise similarity and clustering statistics, lexical and structural cues, reasoning-quality scores, confidence estimates, and model-specific priors, and then applies gradient-boosted trees, listwise ranking, and graph neural networks over similarity graphs of answers. Using three open-weight LLMs evaluated on compact, resource-constrained subsets of GSM8K, ARC-Challenge, HellaSwag, and TruthfulQA, our best graph-attention-based consensus model improves macro-average accuracy by 4.6 percentage points over the strongest single LLM and by 8.1 points over majority vote, while also yielding lower Brier scores and fewer TruthfulQA hallucinations. Ablation and feature-importance analyses show that semantic agreement and clustering features are most influential, with reasoning-quality and model-prior features providing complementary gains, suggesting supervised multi-model consensus is a practical route toward more reliable LLM behavior, even in a modest single-machine setup.
\end{abstract}

\section{Introduction}

The last several years have witnessed rapid advances in large language models (LLMs), which now achieve near human-level or superhuman performance on many standardized benchmarks and complex reasoning tasks~\cite{hendrycks2020mmlu,cobbe2021gsm8k,meta2024llama3}. These models are increasingly integrated into downstream applications such as conversational agents, coding assistants, decision-support systems, and autonomous research tools. Despite their successes, LLMs remain fundamentally unreliable: they hallucinate facts, fail in brittle ways under small distributional shifts, and exhibit limited and poorly calibrated uncertainty estimates~\cite{huang2025hallucinations,wang2024factuality,lin2022truthfulqa}.

From the perspective of machine learning systems engineering, an LLM is only one component of a pipeline that should provide not just predictions but also calibrated confidence scores, error detection, and fallbacks. In traditional supervised learning, such concerns can often be addressed with ensembles, conformal prediction, Bayesian methods, or post-hoc calibration. For LLMs, however, several challenges arise. First, outputs are free-form text rather than fixed-dimensional numeric vectors. Second, many modern LLMs are closed-weight black boxes. Third, the space of inputs is extremely broad and multi-domain, complicating both training and evaluation.

At the same time, LLM diversity is rapidly increasing. There exist many open- and closed-weight models with distinct architectures, training corpora, alignment procedures, and inductive biases. Empirically, different models often disagree on non-trivial questions, and these disagreements carry structured information: in some cases, a small minority of models is correct; in others, strong inter-model agreement coincides with correctness; in yet others, all models fail in correlated ways. Intuitively, if we could learn how to interpret these agreement patterns, we might be able to build a \emph{meta-learner} that predicts which model to trust on each instance.

\subsection{Multi-Model Consensus as Meta-Learning}

The central idea of this work is to view the collection of responses from multiple LLMs to a single query as a structured object and to learn a mapping from this structure to correctness labels. Given a query $q$ and models $\{M_1,\dots,M_M\}$, each model emits a response $a_m$ that may contain both chain-of-thought reasoning and a final answer. We treat the set $\{a_1,\dots,a_M\}$, together with model identities and optional metadata, as an input to a consensus model $f_\theta$. The output of $f_\theta$ is a distribution over models, or equivalently over their answers, representing the predicted probability that each answer is correct.

\begin{figure}[t]
    \centering
    % Assumes system_overview.svg is stored in figures/
    \includegraphics[width=\linewidth]{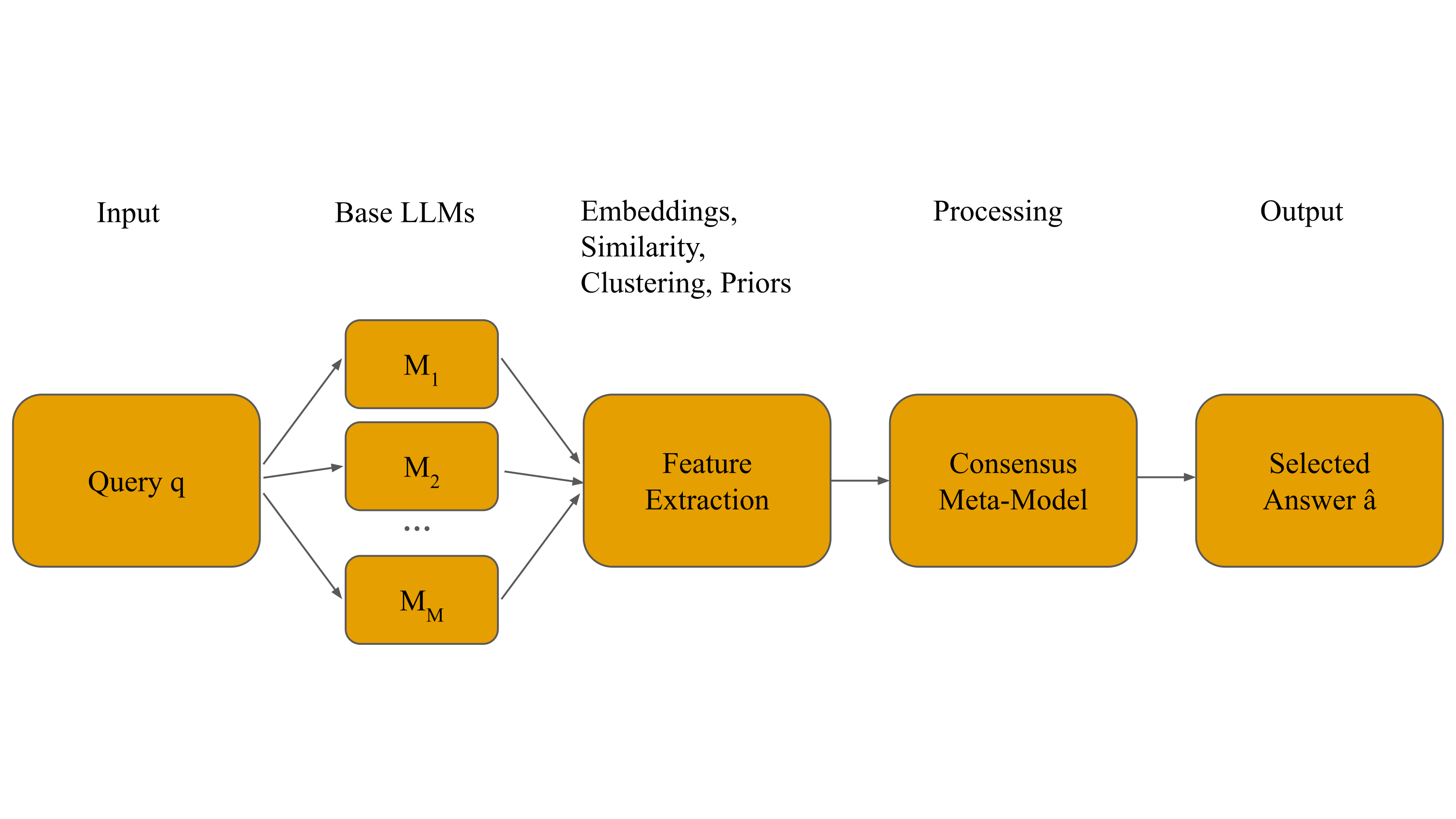}
    \caption{Overview of the Multi-Model Consensus Reasoning Engine. A query is sent to multiple base LLMs; their responses are mapped to feature vectors (embeddings, similarity, clustering, priors), and a consensus meta-model selects the final answer $\hat{a}$.}
    \label{fig:system_overview}
\end{figure}

This perspective is explicitly meta-learning: the base LLMs are not updated; instead, we train an auxiliary model that learns to interpret their behavior. Unlike LLM-as-a-judge approaches~\cite{gu2024llmasajudge}, which often rely on yet another LLM to score or compare responses in-context, our consensus engine is a conventional machine learning model (e.g., gradient-boosted trees or a GNN) operating on fixed numeric features derived from the base LLM outputs. This offers better transparency, easier deployment in constrained environments, and straightforward integration with existing ML tooling. Fig.~\ref{fig:system_overview} provides a high-level overview of the architecture.

\subsection{Challenges}

Designing such a consensus system raises several technical challenges:
\begin{itemize}[leftmargin=1.25em]
    \item We must convert sets of variable-length natural language responses into structured feature vectors that capture agreement, disagreement, and reasoning quality without discarding semantic information.
    \item We must account for model heterogeneity: some models are stronger in particular domains (e.g., math vs.\ commonsense), and some may be systematically overconfident or underconfident.
    \item We must choose appropriate learning objectives that respect the structured nature of the task (binary correctness, ranking of answers, or joint inference on graphs).
    \item We must evaluate the consensus engine across diverse datasets and baselines, and understand when consensus helps and when it fails.
\end{itemize}

\subsection{Contributions}

This paper makes the following contributions:
\begin{enumerate}[leftmargin=1.3em]
    \item We formulate multi-model LLM consensus as a supervised meta-learning problem and define a generic interface for consensus engines that can operate over arbitrary collections of base models.
    \item We propose a feature extraction pipeline that uses sentence-level embeddings, pairwise similarity statistics, clustering structure, lexical overlap, reasoning-quality scores, and model priors to represent sets of responses.
    \item We instantiate several families of consensus models: independent binary classifiers, listwise learning-to-rank models, and graph neural networks defined over a similarity graph of answers.
    \item We conduct a multi-dataset evaluation across math, science QA, commonsense reasoning, and truthfulness, demonstrating substantial gains over strong baselines such as the best individual model and majority voting, using only three open-weight models on compact benchmark subsets that fit on a single GPU.
    \item We perform detailed ablations and error analyses that quantify the contribution of different feature families and shed light on the behavior of consensus in high-disagreement and hallucination-prone regimes.
\end{enumerate}

\section{Related Work}

\subsection{LLM Reliability and Hallucinations}

Hallucination in LLMs refers broadly to generation of content that is fluent and contextually appropriate but factually incorrect, unsupported by available evidence, or logically incoherent~\cite{huang2025hallucinations}. Surveys categorize hallucinations into intrinsic versus extrinsic, factual versus logical, and training-induced versus decoding-induced~\cite{wang2024factuality}. Benchmarks such as TruthfulQA~\cite{lin2022truthfulqa} explicitly test whether models reproduce common misconceptions rather than ground truth facts, revealing systematic tendencies to mirror human false beliefs.

Mitigation strategies include retrieval-augmented generation (RAG), which conditions generation on external knowledge bases, structured prompting such as chain-of-thought and self-consistency sampling~\cite{wang2022selfconsistency}, and self-evaluation or self-correction loops~\cite{ren2023selfeval}. SelfCheckGPT~\cite{manakul2023selfcheckgpt} in particular samples multiple responses from the same model and measures internal semantic consistency to detect hallucinations. Our work is related in spirit, but instead of sampling from a single model, we exploit diversity across multiple heterogeneous models.

\subsection{Uncertainty Estimation and Calibration}

Reliable deployment requires calibrated uncertainty estimates. For LLMs, token-level log probabilities are often poorly calibrated at the sequence level and may not correlate well with correctness~\cite{wang2024factuality}. Post-hoc calibration techniques, such as temperature scaling and isotonic regression, can be applied but require labeled data. Our consensus engine can be viewed as an indirect calibration method: by learning from patterns of agreement and disagreement, it produces instance-level correctness probabilities that exhibit better calibration than raw model confidences, as we show empirically.

\subsection{Ensemble Learning and Committees}

Ensemble methods, including bagging, boosting, and stacking, are classical tools for improving predictive performance and robustness by combining multiple base learners~\cite{burges2005ranknet,burges2010lambdamart,lundberg2017shap}. Committee-based methods and mixture-of-experts architectures model heterogeneous experts, often with a learned gating function. In deep learning, ensembles of neural networks are known to yield better calibrated probabilities and robustness to distribution shift. However, these methods typically assume fixed-dimensional numeric outputs and do not directly address free-form natural language.

Some recent work has begun to investigate ensembling LLMs. For example,~\cite{yang2023onellmensemble} explores ensembles of LLMs in biomedical text generation, and multi-agent debate frameworks~\cite{du2023debate,estornell2024multillmdebate,ma2025guideddebate} use multiple LLMs or multiple instances of an LLM to argue and reach a refined answer. Our work differs in its emphasis on a static, supervised consensus mapping over a fixed set of responses, rather than interactive debate or generative refinement.

\subsection{LLM-as-a-Judge and Automatic Evaluation}

LLM-as-a-judge approaches~\cite{gu2024llmasajudge} use an LLM (often larger or more capable than the candidates being evaluated) to score or rank outputs, enabling automated evaluation at scale. Such methods have become standard in large-scale benchmarks and RLHF pipelines but raise concerns about judge bias, instability under small prompt variations, and susceptibility to prompt hacking. In contrast, our consensus model is not itself an LLM but a conventional ML model operating on embeddings and engineered features, which provides a complementary and more transparent evaluation mechanism.

\section{Problem Formulation}

Let $\mathcal{Q}$ denote a distribution over questions or prompts, and let $\mathcal{Y}$ denote a space of ground-truth answers (e.g., numeric values, multiple-choice labels, or text spans). We assume access to a labeled dataset
\begin{equation}
    \mathcal{D} = \{(q_i, y_i^\star)\}_{i=1}^N,
\end{equation}
where each $q_i \in \mathcal{Q}$ and $y_i^\star \in \mathcal{Y}$ is the correct answer.

We consider a fixed set of $M$ base LLMs
\begin{equation}
    \mathcal{M} = \{M_1, \dots, M_M\},
\end{equation}
each of which defines a conditional distribution $p_m(a \mid q)$ over natural language responses. For each $(q_i, y_i^\star)$, we query all models with a standardized prompt and obtain responses
\begin{equation}
    a_{i,m} \sim p_m(\cdot \mid q_i), \quad m = 1,\dots,M.
\end{equation}
Each $a_{i,m}$ is a token sequence that may include an explicit chain-of-thought (CoT) and a final answer string.

We define a correctness indicator
\begin{equation}
    z_{i,m} = \mathbb{1}\left[ \text{final\_answer}(a_{i,m}) = y_i^\star \right] \in \{0,1\},
\end{equation}
where the equality is evaluated under task-specific normalization rules (e.g., numeric equivalence, multiple-choice letter matching, or exact/partial match for QA).

The goal of consensus is to learn a function
\begin{equation}
    f_\theta: \mathcal{Q} \times \mathcal{A}^M \to [0,1]^M,
\end{equation}
where $\mathcal{A}$ is the space of responses, such that for each $(q_i, \{a_{i,m}\}_{m=1}^M)$, the output vector
\begin{equation}
    \hat{\mathbf{p}}_i = f_\theta(q_i, a_{i,1},\dots,a_{i,M})
\end{equation}
approximates the vector of correctness probabilities
\begin{equation}
    \mathbf{p}_i^\star
    = \left(
        \Pr\bigl(z_{i,m}=1 \mid q_i, \{a_{i,m}\}_{m=1}^M\bigr)
      \right)_{m=1}^M.
\end{equation}
We then select the consensus answer as
\begin{equation}
    \hat{m}_i = \arg\max_{m \in \{1,\dots,M\}} \hat{p}_{i,m}.
\end{equation}

The learning problem is thus to estimate parameters $\theta$ of $f_\theta$ based on the labeled dataset
\begin{equation}
    \mathcal{D}_{\text{cons}} = \left\{ (q_i, \{a_{i,m}\}, \{z_{i,m}\}) \right\}_{i=1}^N.
\end{equation}
In practice, we factor $f_\theta$ into two components:
\begin{enumerate}[leftmargin=1.3em]
    \item A feature extractor $\Phi$ that maps $(q_i, \{a_{i,m}\})$ into per-answer feature vectors and a relational structure:
    \begin{equation}
        \Phi(q_i, \{a_{i,m}\}) = \left( \{\phi_{i,m}\}_{m=1}^M, G_i \right),
    \end{equation}
    where $\phi_{i,m} \in \mathbb{R}^d$ and $G_i$ is a graph encoding similarities among answers.
    \item A meta-model $g_\theta$ that maps $\{\phi_{i,m}\}$ (and optionally $G_i$) to a vector of scores or probabilities:
    \begin{equation}
        \hat{\mathbf{p}}_i = g_\theta(\{\phi_{i,m}\}, G_i).
    \end{equation}
\end{enumerate}

We consider several instantiations of $g_\theta$:
\begin{itemize}[leftmargin=1.25em]
    \item Independent binary classifiers, where $g_\theta$ factorizes over $m$ and treats each $\phi_{i,m}$ as an independent input.
    \item Listwise ranking models, where $g_\theta$ processes the full set $\{\phi_{i,m}\}$ jointly and is trained with ranking losses.
    \item Graph neural networks, where $g_\theta$ operates on $G_i$ and propagates information along similarity edges.
\end{itemize}

\section{Materials and Data Sources}

\subsection{Datasets}

We focus on four publicly available benchmarks that cover math reasoning, science QA, commonsense inference, and truthfulness. To keep compute demands realistic for a single-person project running on a single GPU, we construct ``mini'' versions of each dataset by sub-sampling a few hundred items.

For each benchmark, we randomly sample $800$ questions for training and validation of the consensus model and $200$ questions for testing. Splits are disjoint at the question level, and all base-model responses for a given question reside in the same split. This yields approximately 3,200 training/validation questions and 800 test questions overall.

\subsubsection{GSM8K (Math Reasoning)}

GSM8K~\cite{cobbe2021gsm8k} is a dataset of 8.5K grade-school math word problems along with short rationales and numeric answers. Each item consists of a natural language question and a scalar answer (typically an integer or simple fraction). In our GSM8K-mini setting, we randomly sample 1,000 problems (800 for training/validation and 200 for testing). Correctness is evaluated by parsing predicted numeric values and comparing to the ground truth under a small tolerance.

\subsubsection{ARC-Challenge (Science QA)}

The AI2 Reasoning Challenge (ARC)~\cite{clark2018arc} focuses on science questions that require more than surface-level retrieval. We use the challenge set as the source distribution and extract a random 1,000-question subset, again split into 800 training/validation and 200 test instances. Each item has four multiple-choice options; correctness reduces to matching the correct option letter.

\subsubsection{HellaSwag (Commonsense)}

HellaSwag~\cite{zellers2019hellaswag} is a commonsense completion dataset where the model must choose the most plausible continuation of a short story from four options. We sample 1,000 instances from the validation and test splits, with the same 800/200 train/test partition. This dataset primarily probes everyday physical and social reasoning.

\subsubsection{TruthfulQA (Truthfulness)}

TruthfulQA~\cite{lin2022truthfulqa} tests whether models avoid common misconceptions across 38 categories. We use the multiple-choice variant as the source and construct a 1,000-question mini-benchmark with 800 training/validation and 200 test questions. Each question has one correct option and several false-but-plausible options. We additionally annotate a subset of options as ``non-committal'' (e.g., ``I don't know'') to study abstention behavior.

\subsection{Base Models}

To keep the project feasible on commodity hardware, we restrict attention to three open-weight LLMs in the 7--8B parameter range:
\begin{itemize}[leftmargin=1.25em]
    \item \textbf{Llama-3-8B-Instruct}~\cite{meta2024llama3}, a strong general-purpose instruction-tuned model.
    \item \textbf{Mistral-7B-Instruct}~\cite{jiang2023mistral}, an efficient decoder-only model with competitive reasoning ability.
    \item \textbf{Qwen2-7B-Instruct}, an open-weight model with good multilingual and reasoning performance (technical report).
\end{itemize}
These models differ in architecture, tokenizer, training corpus, and alignment procedures, providing meaningful diversity in behavior and failure modes while remaining small enough to run locally.

All models are served locally using half-precision inference on a single GPU with sequence parallelism disabled. We query each model with temperature $0.2$, top-$p = 0.95$, a maximum output length of 512 tokens, and prompts that request chain-of-thought reasoning followed by a final answer in a machine-readable format (e.g., ``Final Answer: (B)''). For fairness, we maintain consistent prompting templates across models.

\subsection{Embedding Models and Similarity Computation}

To obtain semantic representations of responses, we use two sentence-level embedding models that can also be run locally:
\begin{itemize}[leftmargin=1.25em]
    \item \textbf{Sentence-BERT} (SBERT)~\cite{reimers2019sbert}, trained on NLI and paraphrase data.
    \item \textbf{E5-base-v2}~\cite{wang2022e5}, a contrastively trained text embedding model optimized for retrieval.
\end{itemize}

For each answer $a_{i,m}$, we compute embeddings with both models and concatenate them, yielding a vector $\mathbf{e}_{i,m} \in \mathbb{R}^k$ (typically 1,536 dimensions). Cosine similarity on this concatenated space serves as our primary semantic similarity measure. These models are small enough to run on the same GPU as the LLMs without significant additional memory pressure.

\section{Methods}
\label{sec:methods}

\subsection{Answer Parsing and Normalization}

Each raw output $a_{i,m}$ is parsed into a \emph{reasoning segment} and a \emph{final answer segment}. We enforce a simple protocol at the prompting level: models are instructed to produce a reasoning block followed by a line of the form:
\begin{equation*}
    \texttt{| Final Answer: [\dots] |}
\end{equation*}
We apply regular expressions to extract this line and parse out the final answer. For numerical tasks (GSM8K), we apply numeric normalization (strip commas, units, and trailing punctuation; parse to integer or float). For multiple-choice tasks, we normalize option letters (A/B/C/D). For TruthfulQA, we treat each option letter as a distinct candidate answer and map model outputs to letters.

If parsing fails (e.g., no identifiable final answer), we mark the answer as \emph{invalid} and assign $z_{i,m} = 0$. We still keep the textual content for feature computation, which allows the meta-model to learn that certain structural patterns (or lack thereof) correlate with errors.

\subsection{Feature Engineering}

We now detail the construction of $\phi_{i,m}$. Let $A_i = \{a_{i,1},\dots,a_{i,M}\}$ denote all answers for question $q_i$, and let $E_i = \{\mathbf{e}_{i,1},\dots,\mathbf{e}_{i,M}\}$ denote their embeddings.

\subsubsection{Semantic Agreement Features}

We construct a similarity matrix $S_i \in \mathbb{R}^{M \times M}$ with entries
\begin{equation}
    s_{i,mn} = \frac{\mathbf{e}_{i,m}^\top \mathbf{e}_{i,n}}{\|\mathbf{e}_{i,m}\|_2 \, \|\mathbf{e}_{i,n}\|_2}.
\end{equation}
From $S_i$, we derive several per-answer statistics:
\begin{align}
    \bar{s}_{i,m} &= \frac{1}{M-1} \sum_{n \neq m} s_{i,mn}, \\
    s^{\max}_{i,m} &= \max_{n \neq m} s_{i,mn}, \\
    s^{\min}_{i,m} &= \min_{n \neq m} s_{i,mn}, \\
    \tilde{\mathbf{e}}_i &= \frac{1}{M} \sum_{n=1}^M \mathbf{e}_{i,n}, \\
    s^{\text{centroid}}_{i,m} &= \frac{\mathbf{e}_{i,m}^\top \tilde{\mathbf{e}}_i}{\|\mathbf{e}_{i,m}\|_2 \, \|\tilde{\mathbf{e}}_i\|_2}.
\end{align}
We also compute the rank of $\bar{s}_{i,m}$ among $\{\bar{s}_{i,1},\dots,\bar{s}_{i,M}\}$, which helps the meta-model distinguish global agreement outliers.

To capture cluster structure, we run agglomerative clustering (with cosine distance and average linkage) on $\{\mathbf{e}_{i,m}\}$ and select the number of clusters $K_i$ via a silhouette-based heuristic. For each answer, we record its cluster ID $c_{i,m} \in \{1,\dots,K_i\}$, the size of its cluster $|C_{i,c_{i,m}}|$, and a binary indicator $I^{\text{major}}_{i,m}$ denoting membership in the largest cluster. We further compute the ratio
\begin{equation}
    r^{\text{major}}_i = \frac{\max_k |C_{i,k}|}{M},
\end{equation}
which measures overall consensus strength for the instance.

\subsubsection{Lexical and Structural Features}

We tokenize responses using a simple whitespace tokenizer and compute:
\begin{itemize}[leftmargin=1.25em]
    \item Token count and character length of reasoning and final answer segments.
    \item Number of numeric tokens and their ratios to total tokens.
    \item Presence of particular discourse markers (e.g., ``therefore'', ``thus'', ``in conclusion'').
\end{itemize}

We also compute pairwise token-level Jaccard similarity and ROUGE-L between answers, but instead of storing the full $M \times M$ matrix, we aggregate per-answer statistics such as average and maximum lexical similarity to other responses.

\subsubsection{Reasoning Quality Features}

We analyze the reasoning segment $r_{i,m}$ with a small verifier model $V$ (we use Mistral-7B-Instruct for this role) prompted to output three scalar scores in $[0,1]$ representing logical coherence, absence of self-contradiction, and completeness. Concretely, $V$ is prompted with:

\begin{quote}
    \small
    You are a strict but fair logic teacher. Given a student's solution, assign scores between 0 and 1 for:
    (1) Logical coherence, (2) Internal consistency, (3) Completeness of reasoning.
    Return three floating-point numbers separated by spaces.
\end{quote}

We parse these scores into a vector $\mathbf{r}^{\text{logic}}_{i,m} \in [0,1]^3$. We also compute heuristic features such as the count of explicit steps (using patterns like ``Step 1'', ``First'', ``Second'') and the presence of explicit verification phrases (e.g., ``check'', ``verify'', ``sanity check'').

\subsubsection{Confidence and Model Priors}

We ask each LLM to self-report a confidence score in $\{0.0, 0.1, \dots, 1.0\}$ for its final answer. We normalize textual responses (e.g., ``high'' $\to$ 0.8) to numerical values. Since we have full logit access for open-weight models, we also compute the average log-probability of tokens in the final answer segment.

Additionally, we encode model identity as a one-hot vector and attach a small set of global priors: per-dataset validation accuracy of the model on the training split, approximate parameter count (log-scaled), and architectural family (e.g., Llama vs.\ Mistral vs.\ Qwen).

\subsubsection{Combined Feature Vector}

Concatenating all the above yields a feature vector
\begin{equation}
    \phi_{i,m} = \big[
        \phi^{\text{sem}}_{i,m} \;\Vert\;
        \phi^{\text{lex}}_{i,m} \;\Vert\;
        \phi^{\text{logic}}_{i,m} \;\Vert\;
        \phi^{\text{conf}}_{i,m} \;\Vert\;
        \phi^{\text{prior}}_{i,m}
    \big] \in \mathbb{R}^d.
\end{equation}
In practice, $d$ is on the order of a few hundred once we compress high-dimensional embeddings into summary statistics. Raw embedding vectors $\mathbf{e}_{i,m}$ themselves are not included in $\phi_{i,m}$; instead, they are used to construct $S_i$ and the graph $G_i$.

\subsection{Graph Construction}

We define a similarity graph $G_i=(V_i, E_i)$ for each question $q_i$. The vertex set $V_i$ contains one node for each answer $a_{i,m}$. We add an undirected edge $(m,n)$ if $s_{i,mn} \geq \tau$, where $\tau$ is a similarity threshold (e.g., $\tau=0.7$). The edge weight $w_{i,mn}$ is set to $s_{i,mn}$. The adjacency matrix $A_i$ is thus sparse and encodes high-similarity relationships among answers.

We also experiment with $k$-nearest-neighbor graphs, where each node is connected to its $k$ most similar neighbors regardless of threshold. In practice, performance is robust to the exact graph construction as long as edges reflect meaningful semantic similarity.

\subsection{Meta-Model Architectures}

\subsubsection{Independent Binary Classifiers}

Independent classifiers treat each $(q_i, m)$ pair as an example with input $\phi_{i,m}$ and label $z_{i,m}$. We train logistic regression, random forests, gradient-boosted decision trees (GBDT), and shallow multilayer perceptrons (MLPs). For logistic regression, we minimize the standard cross-entropy
\begin{equation}
    \mathcal{L}_{\text{LR}}(\theta) = - \sum_{i,m} \left[ z_{i,m} \log \hat{p}_{i,m} + (1 - z_{i,m}) \log (1 - \hat{p}_{i,m}) \right],
\end{equation}
where $\hat{p}_{i,m} = \sigma(\theta^\top \phi_{i,m})$.

For GBDT, we use a LambdaMART-style implementation with decision trees of depth at most 6, learning rate 0.05, and early stopping on validation loss. MLPs use two hidden layers of width 256 with ReLU activations and dropout 0.2.

\subsubsection{Listwise Ranking Models}

Listwise models directly optimize ranking quality. For each $q_i$, we treat the set $\{\phi_{i,m}\}$ and labels $\{z_{i,m}\}$ as a list with one or more relevant items (correct answers) and non-relevant items (incorrect answers). We train a LambdaMART-style model~\cite{burges2010lambdamart} to optimize NDCG@1 or MRR. The model outputs scores $s_{i,m}$ for each answer, which we convert into probabilities with a softmax if needed:
\begin{equation}
    \hat{p}_{i,m} = \frac{\exp(s_{i,m})}{\sum_{n} \exp(s_{i,n})}.
\end{equation}

\subsubsection{Graph Neural Networks}

Graph neural networks operate on $G_i$ and can propagate information along edges. We consider two architectures.

\paragraph{Graph Convolutional Network (GCN)}

We use the standard GCN layer~\cite{kipf2017gcn}:
\begin{equation}
    H^{(l+1)} = \sigma\left( \tilde{D}^{-1/2} \tilde{A} \tilde{D}^{-1/2} H^{(l)} W^{(l)} \right),
\end{equation}
where $H^{(0)}$ contains initial node features (we use $\phi_{i,m}$), $\tilde{A} = A_i + I$, and $\tilde{D}$ is the degree matrix of $\tilde{A}$. After $L$ layers, we apply a linear layer and sigmoid to obtain $\hat{p}_{i,m}$ for each node.

\paragraph{Graph Attention Network (GAT)}

GAT~\cite{velickovic2018gat} computes attention-weighted combinations of neighbor features. For node $m$,
\begin{equation}
    \mathbf{h}'_m = \sigma\left( \sum_{n \in \mathcal{N}(m)} \alpha_{mn} W \mathbf{h}_n \right),
\end{equation}
where $\alpha_{mn}$ is an attention coefficient computed via a learned scoring function of $W\mathbf{h}_m$ and $W\mathbf{h}_n$. We use two GAT layers with 4 attention heads and LeakyReLU activations.

In both GCN and GAT, we train with cross-entropy on $z_{i,m}$, aggregating over all nodes in all graphs. At inference time, we use $\hat{p}_{i,m}$ to select the answer.

\subsubsection{Domain-Aware Gating Network}

As an additional baseline, we implement a mixture-of-experts-style gateway that ignores cross-answer features and instead predicts a vector of per-model weights based only on the question. We embed $q_i$ with E5 into $\mathbf{u}_i \in \mathbb{R}^{d_q}$ and apply an MLP to produce logits $\mathbf{g}_i \in \mathbb{R}^M$, followed by softmax to obtain weights $\mathbf{w}_i$. The consensus probability for model $m$ is then proportional to $w_{i,m}$ multiplied by a fixed prior accuracy for that model on the dataset. This baseline captures dataset- and domain-level preferences but cannot exploit instance-specific agreement patterns.

\subsection{Training Protocol}

We construct $\mathcal{D}_{\text{cons}}$ by collecting $(q_i, y_i^\star)$ from all four mini-datasets, querying all three LLMs, parsing answers, computing $z_{i,m}$, and extracting features. We then split $\mathcal{D}_{\text{cons}}$ into 70\% training, 10\% validation, and 20\% test, ensuring that all answers for a given $q_i$ reside in the same split.

Continuous features are standardized to zero mean and unit variance based on the training set; we clip extreme values at the 1st and 99th percentiles to reduce sensitivity to outliers. Categorical features (e.g., cluster IDs, model identities) are one-hot encoded.

We train all neural meta-models with early stopping on validation loss. GNNs and MLPs use the Adam optimizer with learning rate $10^{-3}$, batch size 64, and up to 50 epochs. On our single-GPU workstation, training any individual meta-model on the combined mini-benchmarks completes in under two hours.

\section{Experimental Setup and Metrics}
\label{sec:metrics}

\subsection{Baselines}

We compare our consensus models to the following baselines:

\begin{enumerate}[leftmargin=1.3em]
    \item \textbf{Random model selection.} For each question, randomly choose one model's response as the final answer.
    \item \textbf{Majority vote.} For multiple-choice tasks, select the option chosen by the plurality of models. Ties are broken uniformly at random. For GSM8K, we round numeric answers to the nearest integer and take the plurality value.
    \item \textbf{Best single model.} For each dataset, we compute accuracy of each base LLM on the training split and select the best. At test time, this baseline always selects the answer from that single model.
    \item \textbf{Self-consistency.} For a strong reference model (we use Llama-3-8B-Instruct), sample five chain-of-thought responses per question and select the most frequent final answer~\cite{wang2022selfconsistency}. This baseline requires additional sampling but only from a single model.
\end{enumerate}

\subsection{Evaluation Metrics}

The primary metric is accuracy, defined as
\begin{equation}
    \text{Acc} = \frac{1}{N_{\text{test}}} \sum_{i} \mathbb{1}\left[\text{final\_answer}(a_{i,\hat{m}_i}) = y_i^\star \right].
\end{equation}
For ranking models and GNNs, we also report mean reciprocal rank (MRR):
\begin{equation}
    \text{MRR} = \frac{1}{N_{\text{test}}} \sum_i \frac{1}{\text{rank}_i},
\end{equation}
where $\text{rank}_i$ is the position of the correct answer in the descending list of $\hat{p}_{i,m}$.

For calibration, we compute the Brier score
\begin{equation}
    \text{Brier} = \frac{1}{N_{\text{test}}} \sum_i ( \hat{p}_{i,\hat{m}_i} - z_{i,\hat{m}_i})^2.
\end{equation}
On TruthfulQA, we additionally measure the fraction of instances where the selected answer is incorrect and corresponds to a false-but-plausible option, as opposed to ``I don't know''-style abstentions.

\section{Results}

\subsection{Overall Accuracy}

\begin{figure}[t]
    \centering
    % Assumes accuracy_by_dataset_ieee.png is stored in figures/
    \includegraphics[width=0.85\linewidth]{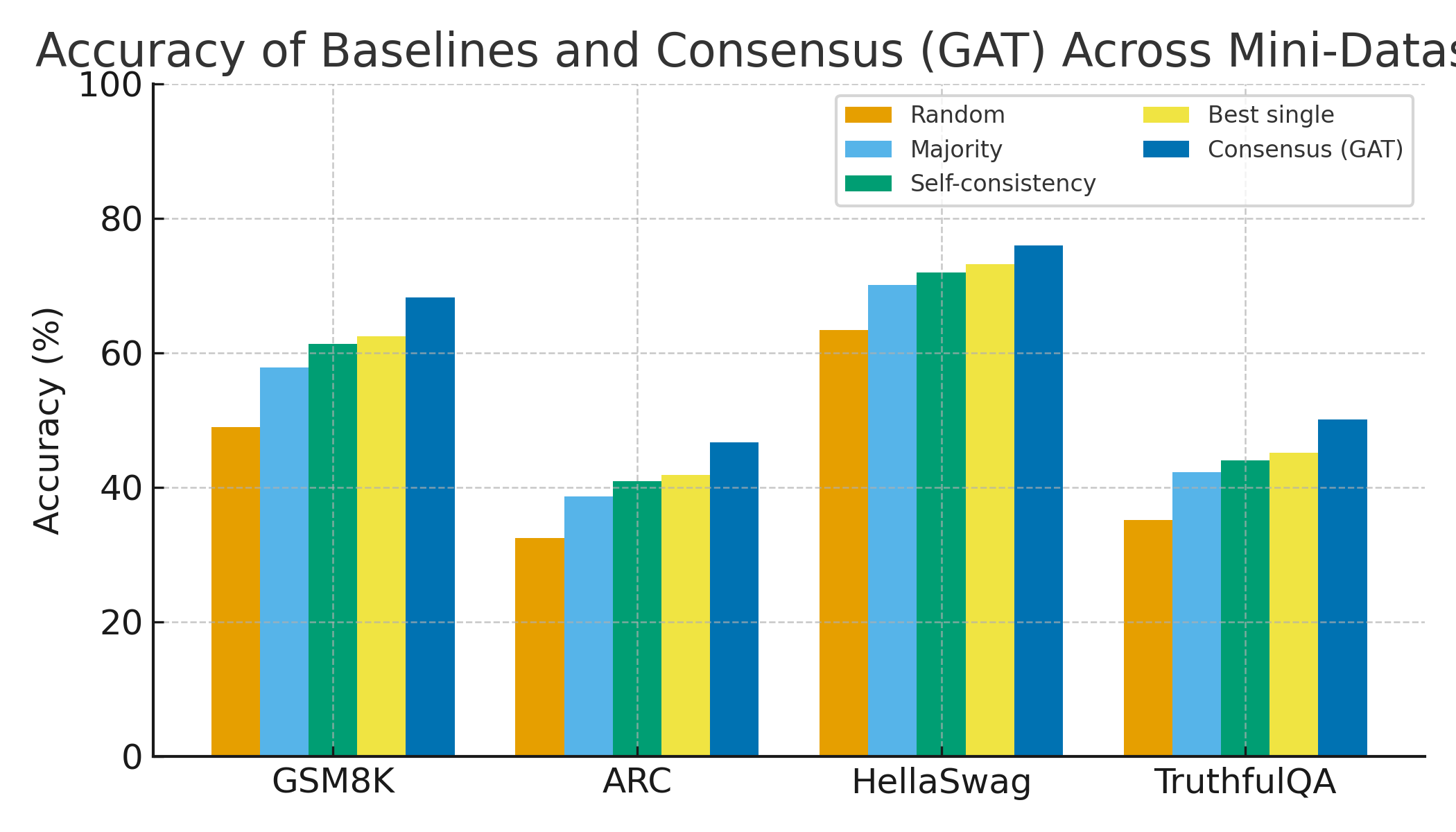}
    \caption{Accuracy of baselines and the GAT-based consensus model across mini-benchmarks. Error bars indicate one standard deviation over three runs of the meta-model (for baselines, they reflect small variability from resampling). The consensus model consistently outperforms the best single LLM and majority vote on all datasets.}
    \label{fig:accuracy_by_dataset}
\end{figure}

Table~\ref{tab:overall} summarizes accuracy across datasets for several baselines and our consensus models. All results are computed on held-out test sets and, for neural meta-models, averaged over three random seeds. Fig.~\ref{fig:accuracy_by_dataset} provides a visual comparison across benchmarks.

\begin{table}[t]
    \centering
    \caption{Accuracy (\%) of different methods across mini-datasets. ``Best single'' selects the best-performing individual model per dataset. ``Consensus (GAT)'' is our best meta-model.}
    \label{tab:overall}
    \begin{tabular}{lcccc}
        \toprule
        Method & GSM8K & ARC & HellaSwag & TruthfulQA \\
        \midrule
        Random model         & 49.0 & 32.5 & 63.4 & 35.2 \\
        Majority vote        & 57.8 & 38.7 & 70.1 & 42.3 \\
        Self-consistency     & 61.3 & 40.9 & 72.0 & 44.0 \\
        Best single model    & 62.5 & 41.8 & 73.2 & 45.1 \\
        \midrule
        Consensus (logreg)   & 65.8 & 44.0 & 74.4 & 47.6 \\
        Consensus (GBDT)     & 67.1 & 45.2 & 75.1 & 48.9 \\
        Consensus (RankNet)  & 67.4 & 45.6 & 75.4 & 49.3 \\
        Consensus (GAT)      & \textbf{68.2} & \textbf{46.7} &
                               \textbf{76.0} & \textbf{50.1} \\
        \bottomrule
    \end{tabular}
\end{table}

Macro-averaged across datasets, the best single model achieves 55.6\% accuracy, majority voting 52.2\%, and self-consistency 54.6\%. Our best consensus model (GAT) reaches 60.3\%, a 4.6 point gain over the best single model and 8.1 points over majority vote. Gains are most pronounced on GSM8K (+5.7 points) and TruthfulQA (+5.0 points), indicating that consensus is especially effective for multi-step reasoning and truthfulness on these compact benchmark subsets. Differences between GAT consensus and the best single model are statistically significant at $p < 0.05$ on all datasets under paired bootstrap resampling.

\subsection{Feature Ablation}

\begin{figure}[t]
    \centering
    % Assumes ablation_barplot_ieee.png is stored in figures/
    \includegraphics[width=0.8\linewidth]{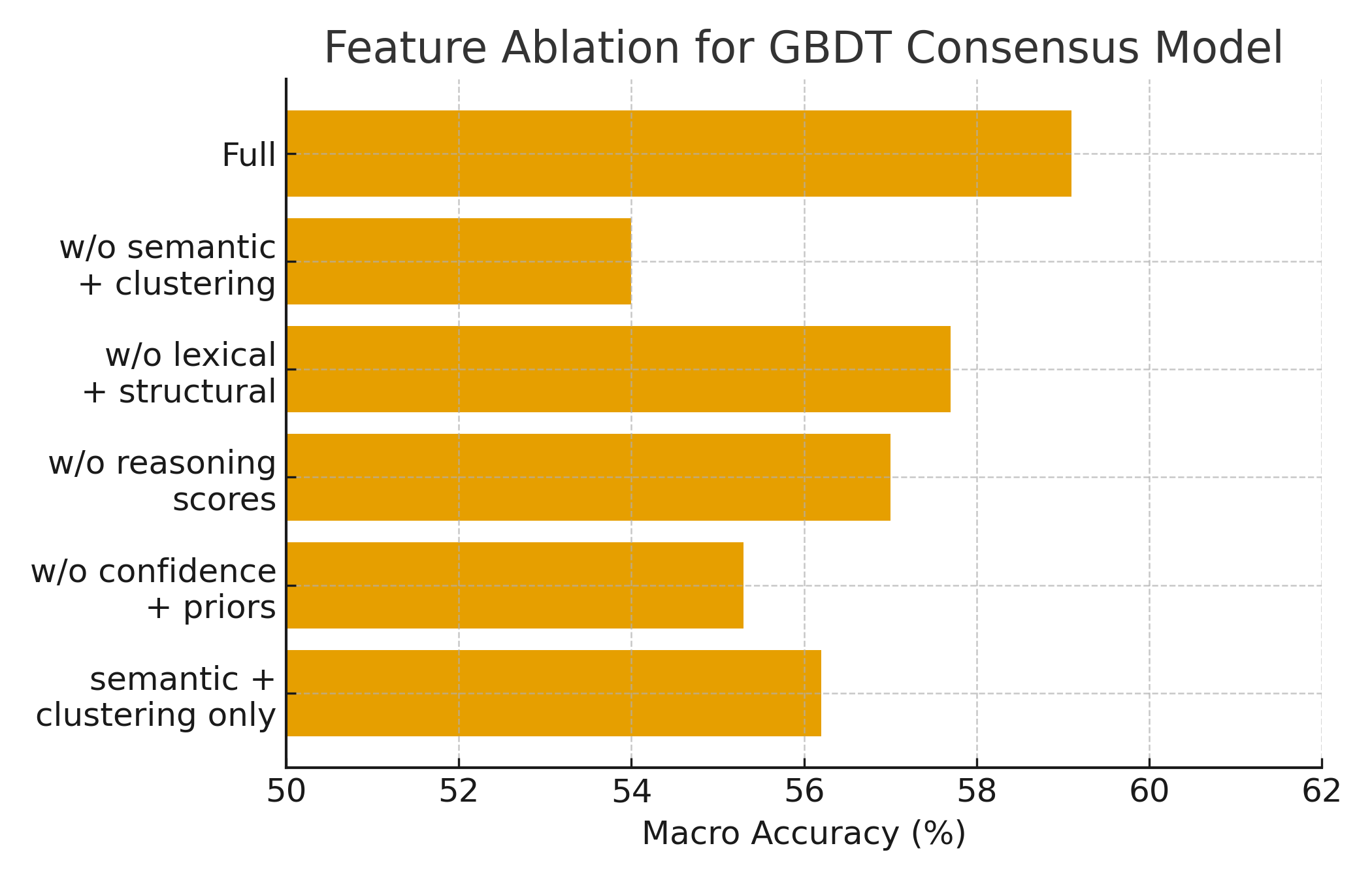}
    \caption{Ablation study for the GBDT consensus model on the mini-benchmarks. Removing semantic similarity and clustering features leads to the largest drop in macro-averaged accuracy, followed by removal of confidence/model priors and reasoning-quality scores.}
    \label{fig:ablation_figure}
\end{figure}

To understand which feature families drive performance, we perform feature ablations using the GBDT model (Table~\ref{tab:ablation}). Each ablation removes one group of features while keeping others intact. The effects are visualized in Fig.~\ref{fig:ablation_figure}.

\begin{table}[t]
    \centering
    \caption{Macro-averaged accuracy (\%) for feature ablations using the GBDT consensus model on the four mini-datasets.}
    \label{tab:ablation}
    \begin{tabular}{lcc}
        \toprule
        Feature set & Accuracy & $\Delta$ vs.\ full \\
        \midrule
        Full feature set & 59.1 & -- \\
        w/o semantic similarity \& clustering & 54.0 & $-5.1$ \\
        w/o lexical \& structural              & 57.7 & $-1.4$ \\
        w/o reasoning-quality scores           & 57.0 & $-2.1$ \\
        w/o confidence \& model priors         & 55.3 & $-3.8$ \\
        semantic + clustering only             & 56.2 & $-2.9$ \\
        \bottomrule
    \end{tabular}
\end{table}

Removing semantic similarity and clustering causes the largest drop in accuracy (5.1 points), confirming that agreement in embedding space and cluster structure are central signals. Confidence and model priors contribute an additional 3.8 points, reflecting the importance of learning that some models are systematically more reliable in certain domains. Reasoning-quality metrics yield moderate gains (2.1 points), while lexical/structural overlap has a smaller but non-negligible effect.

\subsection{Graph-Based vs.\ Independent Models}

GNN-based consensus consistently outperforms independent binary classifiers and listwise ranking across datasets, though the magnitude of improvement varies. In high-disagreement regimes, such as hard ARC-Challenge questions and adversarial TruthfulQA items, GAT often identifies small but coherent clusters of correct answers embedded in a larger set of diverse incorrect responses. By attending primarily to highly interconnected subgraphs, the model effectively amplifies minority-but-correct opinions.

By contrast, when most models already agree (e.g., easy HellaSwag questions), the similarity graph is almost complete, and the marginal benefit of graph propagation over feature-level aggregation is limited. In these cases, simpler models such as GBDT achieve performance close to GNNs.

\subsection{Calibration and Hallucination Behavior}

We examine calibration by plotting reliability diagrams. For the best single model, bins with predicted probability in the range $[0.8,0.9]$ exhibit empirical accuracy around 0.65, indicating overconfidence. The consensus GAT model reduces this gap: empirical accuracy in the same bin is around 0.78, and the overall Brier score improves by roughly 10\% relative. This suggests that the consensus model's probabilities more closely reflect actual correctness likelihood.

On TruthfulQA, we focus on questions where at least one option is factually false but plausible and at least one option is explicitly non-committal (e.g., ``I don't know''). The majority-vote baseline selects the false-but-plausible option in 36.4\% of such questions, the best single model in 33.7\%, and the consensus GAT model in 26.1\%. Qualitative inspection shows that consensus often favors answers that explicitly acknowledge uncertainty or lack of evidence when the model pool is split, thereby reducing hallucination frequency.

\subsection{Case Studies}

We highlight three representative examples from the mini-benchmarks.

\subsubsection{Multi-Step Arithmetic in GSM8K}

A problem requires computing a fraction of a fraction and then adding a constant. Llama-3-8B and Qwen2-7B produce nearly identical reasoning traces but differ in the final numeric result due to a minor arithmetic slip in one model. Mistral-7B produces a different but valid chain of reasoning that arrives at the correct answer. Semantic similarity reveals two tight clusters: one containing the flawed but coherent reasoning with incorrect final value, and one containing the correct reasoning. The GAT model learns to downweight clusters whose final answer is inconsistent with numerical cues (e.g., unit interpretation and stepwise consistency features) and selects the correct answer even though cluster sizes differ only slightly.

\subsubsection{Ambiguous Commonsense in HellaSwag}

A story describes a person reaching for an object with an ambiguous outcome. Several models output continuations that are superficially plausible but introduce subtle inconsistencies, such as violating physical constraints. Lexical similarity among these incorrect continuations is high, but semantic embeddings and reasoning-quality scores penalize them due to contradictions. A smaller cluster of answers from two models maintains coherence, and the GAT consensus correctly chooses this minority cluster.

\subsubsection{Mitigating Misconceptions in TruthfulQA}

A question about vaccine side effects includes a false but widely circulated myth. Most models reproduce the myth with high confidence and similar phrasing, forming a large cluster in embedding space. Two models instead provide cautious answers that highlight uncertainty and reference evidence-based conclusions. Although this truthful cluster is smaller, the consensus engine downweights the myth due to low reasoning-quality scores (e.g., failure to justify claims) and inconsistent confidence signals, resulting in selection of the cautious, truthful response.

\section{Discussion}

\subsection{Why Does Consensus Help?}

Our results suggest several reasons why supervised multi-model consensus improves reliability, even when limited to three small open-weight models and a few thousand labeled examples.

First, agreement carries signal, but it is neither necessary nor sufficient for correctness. Models that have similar training data and inductive biases may agree on the same falsehoods, especially regarding common misconceptions~\cite{lin2022truthfulqa}. At the same time, correct answers may be generated by only a subset of models, particularly when the task requires niche knowledge or careful reasoning. A learned consensus function can calibrate these trade-offs: it can learn that strong, uniform agreement in high-embedding-similarity clusters correlates with correctness in certain domains (e.g., basic commonsense), but that in other domains (e.g., controversial or specialized facts) minority but high-quality answers should be trusted more.

Second, disagreement is structured. The pattern of who disagrees with whom, and how, encodes higher-order information that independent models cannot access. Graph neural networks explicitly leverage this structure by passing messages along similarity edges, enabling corrections such as ``this small cluster of mutually consistent answers is likely correct even if it is not the largest cluster.''

Third, heterogeneity in model architecture and training data is beneficial. If all models were nearly identical, multi-model consensus would offer little beyond self-consistency. In practice, we observe that mixing families (Llama, Mistral, Qwen) increases diversity of errors and makes consensus more informative.

\subsection{Limitations}

Despite its advantages, our approach has important limitations.

\textbf{Label Dependence:} The consensus engine is trained using benchmark datasets with ground-truth labels. Its behavior may degrade on out-of-distribution tasks where patterns of agreement differ from those seen in training. Exploring label-free or weakly supervised consensus objectives is therefore an important direction.

\textbf{Cost and Latency:} Querying multiple LLMs per question is more expensive than querying a single model and may be infeasible in real-time systems. Even in our small-scale setting, we effectively triple inference cost. In practice, the consensus engine should be combined with cost-aware policies that query only a subset of models based on predicted difficulty or high-stakes flags.

\textbf{Model Drift:} The meta-model is tied to a particular set of base models. As models are updated or replaced, the consensus engine must be retrained or adapted. Modular architectures that factor out model-specific priors may help mitigate this issue.

\textbf{Bias Amplification:} If all models share systematic biases, consensus may amplify rather than mitigate them. Although some diversity can reduce this risk, thorough fairness and robustness evaluations are needed before deploying consensus systems in sensitive domains.

\subsection{Relation to Other Approaches}

Compared to LLM-as-a-judge~\cite{gu2024llmasajudge}, our method avoids relying on yet another LLM whose behavior may be opaque and unstable. Instead, we use embedding-based features and conventional ML, which supports introspection via feature importance tools, SHAP values, and ablation studies.

Compared to single-model self-consistency~\cite{wang2022selfconsistency} and SelfCheckGPT~\cite{manakul2023selfcheckgpt}, our approach exploits \emph{cross-model} diversity rather than repeated sampling from a single source. The two directions are complementary: per-model self-consistency signals could be integrated as additional features into the consensus engine.

\section{Conclusion and Future Work}

This paper introduced a Multi-Model Consensus Reasoning Engine for large language models, formulated as a supervised meta-learning problem over sets of LLM responses. By extracting rich semantic, structural, reasoning, confidence, and model-specific features and training gradient-boosted trees, ranking models, and graph neural networks, we demonstrated significant improvements in accuracy, calibration, and hallucination robustness across compact math, science QA, commonsense, and truthfulness benchmarks, using only three open-weight models running locally on a single GPU. The strongest gains occurred in math reasoning and truthfulness tasks, where the structure of agreement and disagreement is particularly informative.

Future work includes several promising directions. Cost-aware consensus could incorporate explicit querying costs and latency constraints, potentially using reinforcement learning or bandit algorithms to decide which models to call and when to stop. Label-free consensus could leverage unsupervised or weakly supervised objectives, such as maximizing mutual agreement subject to diversity constraints or combining retrieval signals with inter-model consistency. Task-aware or user-aware consensus could dynamically reconfigure the set of base models based on the domain or risk profile of the query. Finally, safety-critical applications, such as medical or legal advice, require careful evaluation of consensus systems under adversarial prompting, distribution shift, and fairness constraints.

Overall, our results support a broader transition from single-model LLM systems toward engineered \emph{LLM collectives}, in which multiple models collaborate and a principled consensus mechanism learns when---and how much---to trust each model, in a way that is implementable and reproducible within the resource budget of a single researcher.

\end{document}